\documentclass[lettersize,journal]{IEEEtran}
\usepackage{amsmath,amsfonts}
\usepackage{array}
\usepackage{textcomp}
\usepackage{stfloats}
\usepackage{url}
\usepackage{verbatim}
\usepackage{graphicx}
\usepackage{cite}
\usepackage[ruled,vlined]{algorithm2e}
\usepackage{multirow}
\usepackage{xcolor}
\usepackage{subfigure}
\usepackage[colorlinks,
            linkcolor=red,
            anchorcolor=blue,
            citecolor=green
            ]{hyperref}

\begin{document}

\title{Constrained Multiview Representation for Self-supervised Contrastive Learning}

\author{Siyuan Dai, Kai Ye, Kun Zhao, Ge Cui, Haoteng Tang*, and Liang Zhan*~\IEEEmembership{}
\thanks{Siyuan Dai, Kai Ye, Kun Zhao, Ge Cui, Liang Zhan are with the Department of Electrical and Computer Engineering, University of Pittsburgh, Pittsburgh, PA 15213 USA (e-mail: siyuan.dai@pitt.edu). Haoteng Tang is with the Department of Computer Science, University of Texas Rio Grande Valley, Edinburg, TX  USA (e-mail: haoteng.tang@utrgv.edu). Haoteng Tang and Liang Zhan are corresponding authors}
\thanks{Manuscript received XX, 2024; revised XX, 2024.}}

\markboth{IEEE TRANSACTIONS ON Image Processing, VOL. XX, NO. XX, XXXX 2024}%
{Shell \MakeLowercase{\textit{et al.}}: A Sample Article Using IEEEtran.cls for IEEE Journals}

\maketitle

\begin{abstract}
Representation learning constitutes a pivotal cornerstone in contemporary deep learning paradigms, offering a conduit to elucidate distinctive features within the latent space and interpret the deep models. Nevertheless, the inherent complexity of anatomical patterns and the random nature of lesion distribution in medical image segmentation pose significant challenges to the disentanglement of representations and the understanding of salient features. Methods guided by the maximization of mutual information, particularly within the framework of contrastive learning, have demonstrated remarkable success and superiority in decoupling densely intertwined representations. However, the effectiveness of contrastive learning highly depends on the quality of the positive and negative sample pairs, i.e. the unselected average mutual information among multi-views would obstruct the learning strategy so the selection of the views is vital. In this work, we introduce a novel approach predicated on representation distance-based mutual information (MI) maximization for measuring the significance of different views, aiming at conducting more efficient contrastive learning and representation disentanglement. Additionally, we introduce an MI re-ranking strategy for representation selection, benefiting both the continuous MI estimating and representation significance distance measuring. Specifically, we harness multi-view representations extracted from the frequency domain, re-evaluating their significance based on mutual information across varying frequencies, thereby facilitating a multifaceted contrastive learning approach to bolster semantic comprehension. The efficacy of our proposed framework was rigorously evaluated using publicly accessible CT-captured lung lesion segmentation datasets and compared against numerous influential segmentation models backboned with either pure CNN modules or transformer modules. The statistical results under the five metrics demonstrate that our proposed framework proficiently constrains the MI maximization-driven representation selection and steers the multi-view contrastive learning process. The statistical results under the five metrics also show the superiority against the compared baselines.
\end{abstract}

\begin{IEEEkeywords}
Mutual Information Maximization, Representation disentangle, Contrastive Learning, Frequency domain, Lung Lesion Segmentation.
\end{IEEEkeywords}

\section{Introduction}
\IEEEPARstart{D}{iscovering} useful representations stand as a foundational objective within the deep learning models, and developing effective methodologies for representation extraction, selection, and learning remains a challenging problem. Traditionally, researchers concentrated on modeling a function $g$ as a feature extractor, e.g. encoder, which simultaneously reduces the dimension of the input data and also maps it into a representation-separated feature space\cite{he2016deep}, \cite{simonyan2014very}, \cite{he2020momentum}.
For example, given an input image $x$, we could easily get the feature representation $g(x)$ utilizing the ubiquitously employed pre-trained models, combined with a streamlined head $d$ to accomplish particular downstream tasks like image classification and object detection\cite{he2020momentum}, \cite{ren2015faster}. 
The architecture of the head $d$ is usually designed as linear layers, given that the outputs, i.e. the representations themselves, manifest as low-dimensional vectors. Things come to the same in the natural language processing domain, particularly for word representation generation\cite{mikolov2013distributed}, \cite{pennington2014glove}.

\begin{figure}[t]
\centering
\includegraphics[width=7.5cm]{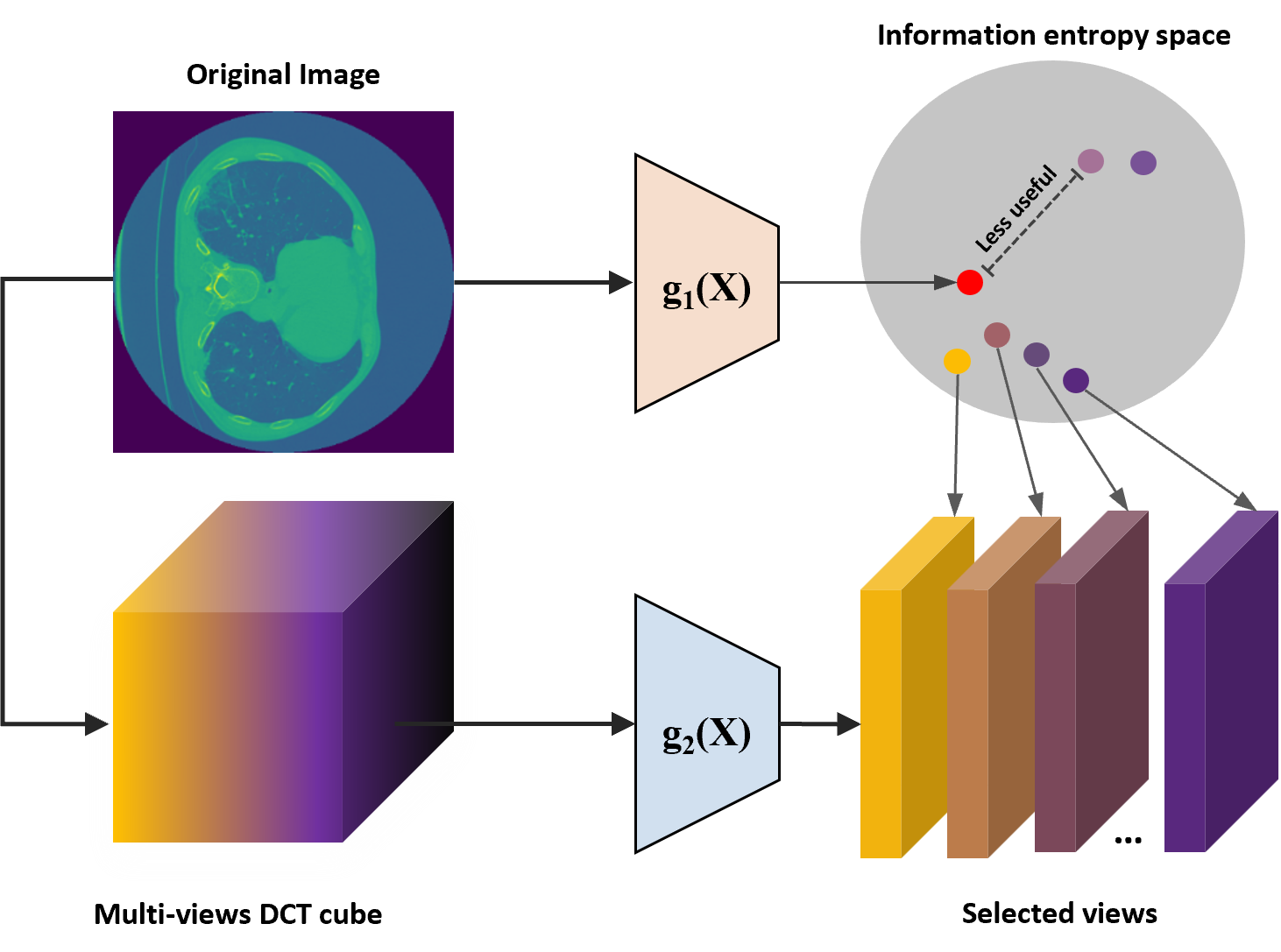}
\caption{Given a generated multi-view DCT cube from the original image, deep representation features will be re-ranked and selected according to the distance with the original image representation in the information entropy space.} 
\label{fig: mi}
\end{figure}

It is inherently understood that the key to analyzing input data pivots on the disentanglement of increasingly representative latent representations denoted as $g(x)$.
Mutual information (MI) is a good tool for quantifying the interdependency between two variables. 
Particularly, consider two variables $X$ and $Y$ are mapped into the same feature space; we may opt to minimize the distance (e.g., $L_{1} norm$, $L_{2} norm$) or amplify the similarity (e.g., cosine similarity) between them. However, this approach is not conducive when dealing with the input image and its low-dimensional representations after being processed by deep neural networks which are constructed with too much nonlinear transformation. 
According to the information theory, let $H(X)$ represent the information entropy of the variable $X$, and $H(X|Y)$ will be the conditional entropy when given the variable $Y$, then $I(X;Y) = H(X)-H(X|Y)$ could amount the information uncertainty between variable $X$ and $Y$ and they could know be analyzed in the same information entropy space.
The higher the MI is, the lower the uncertainty is, and also the stronger the representative is.
Thus, the maximization of the MI, particularly between the extracted representation $g(x)$ and the input $x$, $I(x;g(x))$, facilitates selecting more meaningful representations, which is a typical self-supervised learning strategy\cite{tschannen2019mutual}.

The efficiency of contrastive learning depends on the quality of paired positive and negative representations while meaningful representation extraction leads to an MI-based multi-view variable selection\cite{becker1992self}.
A conventional multi-views generating approach involves sampling different patches from the original $x$ into $\{x^{(1)}, x^{(2)}, ..., x^{(n)}\}$, and they could be possibly overlapping views of $x$. 
Contrastive multi-view coding (CMC)\cite{tian2020contrastive} advocates for obtaining multi-views $x^(i)$ from different image modalities (e.g. different color channels, image segmentation masks). 
Moreover, in contrast to the relatively fixed patterns of objects in general semantic segmentation scenarios\cite{geiger2012we}, \cite{lin2014microsoft}, \cite{mottaghi2014role}, the morphology and distribution of lung lesions are in much more complex and various characters.

To address this challenge, we propose a novel multi-view contrastive learning strategy based on mutual information-guided feature selection within the frequency domain. 
Recent researchers have already taken advantage of the information in the frequency domain. Especially for the application in image compression\cite{gueguen2018faster}, \cite{ehrlich2019deep}, \cite{xu2020learning}, and also for model compression or pruning\cite{chen2016compressing}, \cite{liu2018frequency}, \cite{wang2018packing}. 
Inspired by \cite{qin2021fcanet} from the construction of a multi-frequency attention pipeline, which enhances the effectiveness of the original global average pooling (GAP)\cite{lin2013network} by capturing multi-view attention from disparate frequency channels.
Moreover, different from the paradigm of camera-captured RGB images, CT imaging, when undertaken under varying settings(e.g. routine CT, enhancement CT, low-dose CT, spectral CT), causes semantic instability for identical organ areas across different patients, attributed to the requisite consideration of disparate abdominal structures and radiological information inherent to each patient. 
Transposition of images into the frequency domain could release some obscured features in the original spatial domain\cite{zhong2022detecting}, \cite{cuthill2019camouflage}, \cite{stevens2009animal}.

However, some of the views in the frequency domain are not significant enough for contrastive representation learning according to the compression and pruning works mentioned above. 
Although intuitive MI maximization between views would enhance the effectiveness of representation learning, some works\cite{tschannen2019mutual} argue that the success of these methods cannot be attributed to the properties of MI alone and direct maximization can even result in worse representations, the quality of views affect the performance a lot. 
Especially in medical lesion segmentation tasks, while the output feature map retains the same dimensionality as the input, the semantic divergence necessitates an evaluation of the differences in semantic interpretation or information entropy. 
Previous works\cite{tian2020contrastive}, \cite{hjelm2018learning} grounded in mutual information-based representation learning tried to confine the representation, thereby facilitating the reconstruction of the input and ensuring that the learned representation with minimal uncertainty for conditional information entropy relative to the input.
Concurrently, several representation learning works\cite{rumelhart1985learning}, \cite{ballard1987modular} unrelated to mutual information focus on the quality of reconstructing the original input data. 
This implies an expectation that $d(g(x))=x$, essentially employing the output as a variant of the view\cite{tian2020contrastive}.
\cite{oord2018representation} contemplated the disparities in representation between the input and output, encoding the target $d_{t1}(g(x))$ and context $g_{t2}(x)$ into densely distributed representation vectors, employing non-linear learned mappings within a time-sequence model framework. This strategy is instrumental in preserving low-level features that risk loss during the feature extraction and dimension reduction phases on high-dimensional feature maps.
While $g_t$ represents the different levels of feature extraction and $d_t$ denotes the disparate tiers of representation mapping directed at downstream tasks, all mapping within a time-sequenced architectural framework.
However, these approaches predominantly concentrate on classification tasks, with a core objective of preserving different levels of semantic information.
Although such a method\cite{tian2020contrastive} also integrates the output into the multi-view mutual information estimation process, the classification and depth estimation tasks have a naturally low disparity relatively in the underlying representation between the input and output.

In this study, we focus on the selection of high-quality representations for conducting contrastive representation learning which is particularly suitable for the lung lesion segmentation downstream tasks by continuously maximizing the MI for enhancing the effectiveness of latent representations and selecting top-rank useful views.
Specifically, denoting the output mask as $Y$, we think that it is unwise to assume and expect $H(X) \approx H(g(X)) \approx H(Y)$ and to directly maximize the multi-view mutual information such as $I(X;g(X);Y)$, especially given the evident disparity where $X \ne Y$.
We assume that $H(X)=H(\Bar{X})+H(\Tilde{X})$, and in the segmentation situations, $H(\Bar{X})$ represents the entropy of segmented instances while $H(\Tilde{X})$ goes to the other backgrounds. 
In the classification, the regression-based tasks like object detection and depth estimation, both the $H(\Bar{X})$ and $H(\Tilde{X})$ would directly help to get the output $Y$. 
Although it is much wiser to use more $H(\Bar{X})$ for mask generation, $H(\Bar{X})$ and $H(\Tilde{X})$ have high correlation since the structure information of $Y$ and the spatial information between $H(\Bar{X})$ and $H(\Tilde{X})$ is much stable. 
But in our lung lesion segmentation task, the abdominal structure is fixed but the pathology structure of the lesion is totally random, and also for the spatial information between the background lung and the lesion. 
In our framework, besides learning useful representations under the contrastive strategy, we also continuously maximize and select those more helpful latent features from the multi-view representation in the frequency domain.

Our main contribution to this article is summarized below:
\begin{itemize}
\item{We introduce a new frequency domain-based multi-views generation strategy for self-supervised contrastive learning, which is also easy to expand as semi-supervised learning when getting the mask involved.}
\item{We propose a novel continuous mutual information maximization and score-ranking method for feature selection, solving the problem of preventing those less useful views into contrastive learning.}
\item{We conduct extensive experiments on 3 public lung lesion datasets and compare them with 8 commonly used pure CNN block baselines and 3 influential transformer-based baselines under 4 evaluation metrics to show the superiority of our proposed representation learning frameworks. Furthermore, experiments under different representation learning settings and parameter analysis also demonstrate the effectiveness of our novel method.}
\end{itemize}

\section{Related Work}
Medical image representation learning is a pivotal topic within the domain of computer vision.
In this section, we specifically concentrate on three principal domains that have significant relevance to our proposed framework.
Initially, we will introduce the contemporary representation learning methodologies within the expansive domain, subsequently for the mutual information technique and its applications.
Conclusively, we will talk about the utilization of frequency domain information within related specific domains, demonstrating its potential in providing multi-view images to benefit representation learning.

\subsection{Representation Learning}
The success of machine learning intrinsically depends on the quality of data representations\cite{bengio2013representation}, prompting researchers to focus on foundational models. These models are aimed at learning useful representations for downstream tasks, leading to the ubiquitous use of effective pre-trained model components.
Notably, within the realm of Natural Language Processing (NLP), the mapping of abstract words to vector spaces\cite{mikolov2013distributed,fu20233d,pennington2014glove} has significantly unlocked the potential for efficient learning and processing by neural networks.
\cite{matthew2018peters}, \cite{kenton2019bert} leverage context-sensitive information to encode valuable representations, thereby dominating challenging downstream tasks. 
Similarly, numerous influential works\cite{donahue2014decaf}, \cite{he2016deep}, \cite{he2020momentum}, \cite{carreira2017quo}, \cite{chen2020simple} have achieved state-of-the-art performance across computer vision tasks. However, He et al.\cite{he2019rethinking} thought that the general effectiveness of pre-trained computer vision models warrants further exploration and critical evaluation.
The transferability of representations across different vision tasks within current training datasets presents challenges. It is wise to involve task-specific knowledge during the representation learning process. This approach\cite{bengio2013representation} aids in identifying and disentangling the diverse underlying factors that are either obscured or omitted within the low-level feature maps.
Subsequent works, such as SwAV\cite{caron2020unsupervised} and BYOL\cite{grill2020bootstrap}, have endeavored to enhance the quality of the learned representations. Nevertheless, while these models excel in classification or object detection tasks, their performance was easily beaten on segmentation tasks\cite{xiao2021region}.
Tete et al.\cite{xiao2021region} have demonstrated that task-specific representations necessitate a focus on varied feature levels. Classification tasks are more suitable for global image-level information, whereas segmentation tasks demand an emphasis on localization optimization.
In our work, we prevent force optimizing the network for either global or local representations, acknowledging the stochastic nature of pathology structures and distribution information. 
We re-rank the significance of various representations obtained from multi-view perspectives in the frequency domain and select the top useful views, ensuring a comprehensive and nuanced understanding of the data.

\subsection{Mutual Information}
The main objective of machine learning is to wisely extract and utilize information for downstream practical tasks.
In contrast to the strategy of deep metric learning, which involves mapping features into a designated deep space and measuring under geometric concepts for capturing the meaningful information\cite{hoe2021one}, \cite{kaya2019deep}, \cite{lim2022hypergraph}, \cite{cakir2019deep}.
Mutual information is a concept in information theory measured with probability statistics and serves to estimate the dependency between two random variables.
It is also a useful tool for calculating the relationship between the information entropy encompassed within two different information instances under the same information space.
Predicated on this attribute, the application of mutual information in neural network representation learning has witnessed broad success, originating from the \textit{Infomax} optimization principle\cite{linsker1988self}, \cite{bell1995information}.
This principle advocates the maximization of MI between the input and output of a neural network, despite the notorious challenge posed by estimating MI in contemporary deep networks.
Belghazi et al\cite{belghazi2018mine} proposed the MINE algorithm, an innovative approach for estimating the MI of continuous variables. We have also embraced this method to estimate the MI in our work.
Furthermore, some researchers have delved into various methodologies for calculating MI within deep neural networks, notably modifying the MI objective function based on the Jensen-Shannon divergence for their Deep \textit{InfoMax} algorithm\cite{hjelm2018learning}, and Ozair et al\cite{ozair2019wasserstein} considered the MI estimation based on the Wasserstein divergence.
With the MI maximization methods being measured, some downstream applications spread.
Sanchez et al.\cite{sanchez2020learning} employed a foundational method for representation disentangling, unraveling the representation into a shared part and an exclusive part, thereby achieving new superior classification state-of-the-art results. 
Shuai et al.\cite{zhao2019region} introduced a novel loss function aimed at maximizing the MI between regions. This approach compels the network to more straightforwardly and efficiently concentrate on pixel interdependencies, thus enhancing the effectiveness of semantic segmentation tasks.
Region MI pushes the network to focus on diverse image perspectives, with an increasing number of researchers acknowledging the importance of multi-view analyses.
\cite{bachman2019learning}, \cite{tian2020contrastive}, \cite{oord2018representation} also proposed novel methods for calculating the MI from different views and achieving awesome results. 
In our works, we have also put forward a novel strategy for providing multi-view information, mapping the original data into the frequency domain.
Furthermore, inspired by \cite{covert2023learning}, we selectively consider views with high correlation for the purpose of more effective contrastive learning.

\subsection{Frequency Domain Information}
Traditional computer vision algorithms predominantly focus on image analysis within the spatial domain, specifically \textit{RGB} or \textit{Gray-Scale} images, which align seamlessly with human visual perception.
However, the information within this domain may mask intricate details. Research\cite{cuthill2019camouflage,stevens2009animal} has elucidated that in processing visual scenes, animals own broader wavebands compared to humans, attributed to their proficiency in identifying features from the frequency domain.
Consequently, some researchers advocate that leveraging the Discrete Cosine Transform (DCT) to transpose original images into the frequency domain serves as an effective method for compressing not only the images\cite{gueguen2018faster,ehrlich2019deep,xu2020learning} but also the networks architecture themselves\cite{chen2016compressing,liu2018frequency,wang2016cnnpack}.
Information within the frequency domain can be harnessed for compression purposes, as salient semantic information is easier to extract in this particular feature space.
The attention mechanism forces networks to concentrate on the most crucial parts of the feature maps so utilizing frequency information for formulating attention pipelines is considered.
Qin, et. al\cite{qin2021fcanet} found that global average pooling (GAP) is a commonly used method for designing channel attention mechanisms.
A potential weakness arises when disparate channels share identical mean attention values, from such channels harbor divergent levels of semantic information, leading to bias with attention information.
They claimed that Global Average Pooling (GAP) represents a special instance of DCT, similar to its mean representation across the full frequency domain, thereby advocating a multi-frequency channel attention framework.
Additionally, the FSDR\cite{huang2021fsdr} architects an innovative attention pipeline grounded on frequency space information to improve the domain generalizability for networks, compelling the network to learn features less influenced by specific domains and more aligned with intrinsic semantic features.
Furthermore, the frequency domain can significantly aid in low-level vision scene tasks\cite{liu2018multi,zheng2021learning}, employing the DCT or Wavelet transformation for enhanced performance.
Taking into account these previous works, the frequency domain emerges with good potential for providing multi-view information, conducive to contrastive representation learning.

\section{Methodology}
Our goal in this work is to learn useful representations by leveraging multi-view data with minimal uncertainty for contrastive learning.
We will begin with the generation of multi-views by transferring the original data into the frequency domain.
Subsequently, some preliminary knowledge about information theory and mutual information will be introduced, illustrating the processes of feature selection and the foundation theory of this module.
Thereafter, contrastive learning with multi-view representations will be conducted, proposing the contrastive loss from the single-view situation to the multi-view paradigm. 
Notably, our contrastive learning is conducted under a self-supervised setting which means the representation process will not involve the ground-truth mask, it is easy to be a semi-supervised contrastive learning if the segmentation mask is involved. 
Finally, the segmentation framework with a detailed optimization pipeline will be illustrated.

\subsection{Muti-view Data Generation In The Frequency Domain}
\begin{figure}[ht]
\centering
\includegraphics[width=8cm]{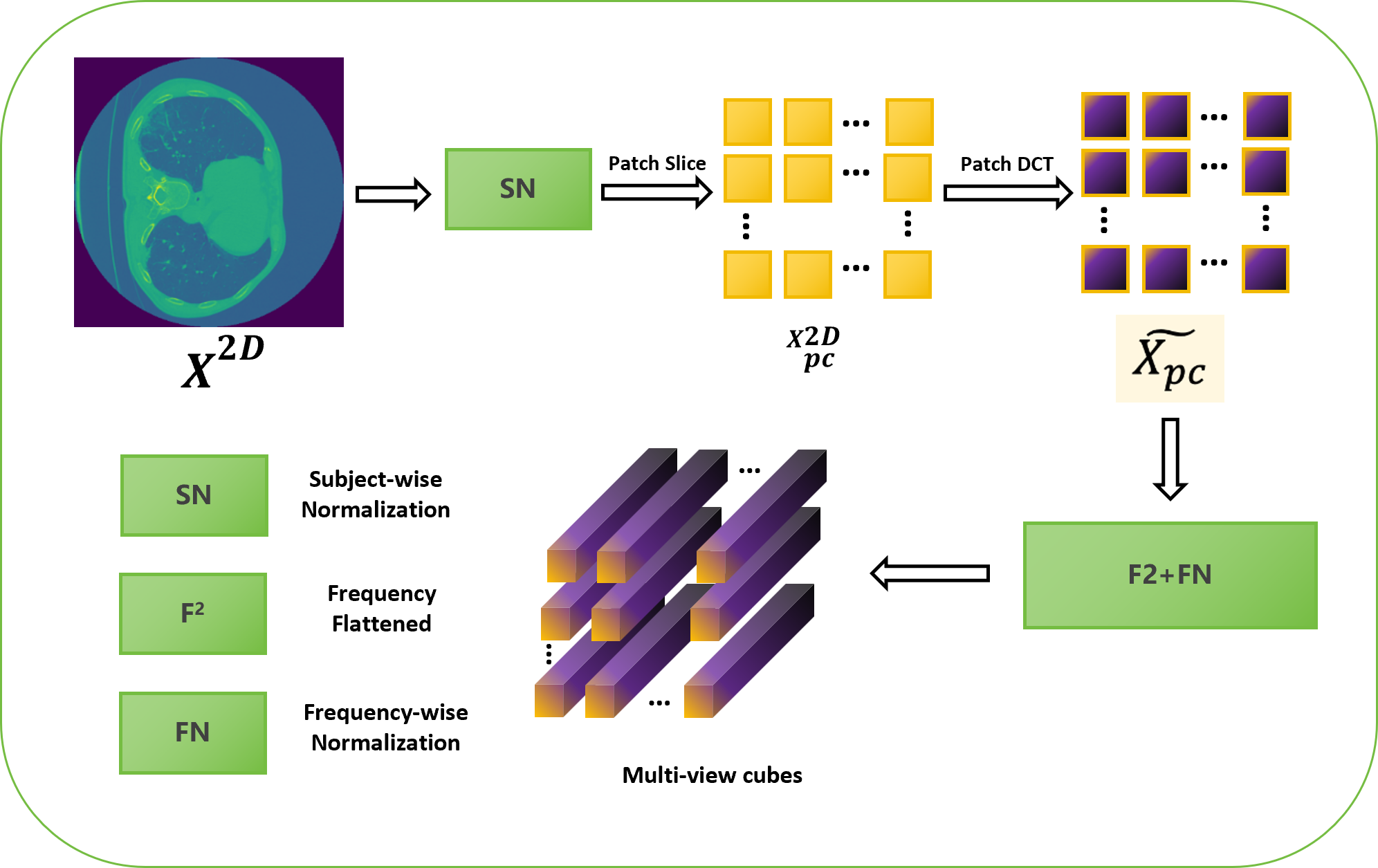}
\caption{An overview of the off-line DCT transformation module for generating multi-view cube. Every original image slice is partitioned into small image patches after subject-wise normalization. Consequently, a DCT transformation is implemented on image patches. Finally, the coefficient cube for the whole image is generated from frequency-based flattened ($F^{2}$) and frequency-wise normalization (FN) operations.} 
\label{fig:dct}
\end{figure}

In our work with CT images, they are naturally captured as a 3D volume but not like general RGB images, so we cannot simply presume the intensity scale is all in $[0, 255]$ and the intensity in the images captured from different institutions and devices would also be different, so normalize the original data into the same intensity scale is necessary. 
We normalize both the liver CT one patient by one patient with the min-max value of themselves, conducting subject-wise min-max normalization (SN) after moving out the slices without any lesions, mapping them to the same intensity scale in the spatial domain.
After normalization, we slice all the 3D CT volumes into 2D images (i.e., $X^{2D}\in{\mathcal{R}^{{H}\times{W}}}$, where $H$ and $W$ denote image size) are generated, we conduct DCT transformation to convert them into the spectrum domain for another modality (see Figure \ref{fig:dct}). 
Particularly, the DCT transformation is conducted patches by patches (We set the patch size as $8 \times 8$ in default, and experiments under different sizes will also be demonstrated) on \textit{2D} images, to extract more fine-grained features in the frequency domain. 
The DCT transformation (i.e., $\hat{X} \in \mathcal{R}^{1 \times 8 \times 8}$) on every \textit{2D} image patch is computed as follows:

\begin{equation}
    \begin{aligned}
        &\hat{X}\left ( i,~j\right )=\frac{2}{\sqrt{\left ( N_1,~N_2 \right )}}\sum_{n_1=0}^{N_1-1}\sum_{n_2=0}^{N_2-1}X^{2D}\left ( n_1,~n_2 \right )\cdot \\
        &a_{n1}a_{n2}cos\left [ n_1\frac{2\pi }{n_1}(n_1+\frac{1}{2}) \right ]cos\left [n_2\frac{2\pi}{N_2}(n_2+\frac{1}{2})\right ] \\
            &s.~t.~a_{n1},~a_{n2}=
            \left\{\begin{matrix}
                \frac{1}{\sqrt{2}},&~k=0\\ 
                1,~&k\neq 0 
            \end{matrix}\right.    
    \end{aligned}
    \label{equa:dct}
\end{equation}

When $i, j, n_1, n_2$ are in range of $[0, 7]$ by default, the $N_1=N_2=8$, $a_{n1},~a_{n2}$ are the constant coefficient. 
To collect multi-view frequency feature maps along $2D$ images and patches, the $\hat{X}$ is flattened according to the frequency ($F^{2}$) from $1 \times 8 \times 8$ to the size of $64 \times 1 \times 1$, while the first number represents the channel and the last two refer as the length and the width, and every channel refers as the feature in different frequency under the spectrum space. 
We then group the feature maps from all image patches and generate the channel-wise DCT coefficient cube as $ \hat{X} \in \mathcal{R}^{64 \times H/8 \times W/8} $.

Since the intensity value after DCT transformation would be mapped to a high range of scale in different frequency feature representations it is difficult for neural networks to handle and extract useful representations.
We then implement another frequency-wise normalization (FN) channel by channel for every DCT coefficient cube and let them in the range of $[0, 1]$.

\subsection{Constrained Multi-view Mutual Information}
\subsubsection{Mutual Information}
As we mentioned in the first section, assume the inputs as a variable $\{X|X=(x_1, x_2, ..., x_I)\}$ and also an entity in the information entropy space(both useful and useless information are involved) while every lowercase $x$ denotes different image instances.
Cause $X$ is discrete, we use its probability mass function (PMF) $p(x)$ for measuring the information entropy, and it can be defined as Equation \ref{eq:entropy} according to \cite{cover1999elements}:

\begin{equation}
\label{eq:entropy}
    H(X)=-\sum_{x \in X} p(x) \log p(x).
\end{equation}

Since entropy can measure the uncertainty of a variable. We can extend such a concept to conditional entropy for representing the amount of uncertainty remaining about one random variable given the knowledge of another. Precisely, conditional entropy can be mathematically articulated as in Equation \ref{eq:condition_entropy}.
This formulation inherently relates to the joint entropy of the two variables, which encapsulates the total uncertainty inherent in the joint distribution. It is confident since it delineates the extent to which the information of one variable reduces uncertainty about the other.

\begin{equation}
\label{eq:condition_entropy}
    H(X \mid Y)=-\sum_{y \in {Y}} {\sum_{x \in X}} p(x, y) \log p(x \mid y).
\end{equation}

Leveraging the principles of information theory, mutual information quantifies the dependency between variables, delineating the uncertainty reduction in variable $Y$ owing to variable $X$, as expounded by Hyvärinen et al\cite{hyvarinen2000independent}. 
Within neural networks, this concept translates to defining $Y$ as the feature map in latent space, representing the learned, essential representation of input data.
This dependency is quantitatively captured in Equation \ref{eq:mi}, reflecting the mutual information between the latent space representation and the input data. 

\begin{equation}
\label{eq:mi}
\begin{aligned}
    I(X;Y) &=H(X)-H(X \mid Y) \\
        &=\sum_{y \in {Y}} \sum_{x \in {X}} p(x, y) \log \frac{p(x, y)}{p(x) p(y)} \\ 
        &=\sum_{g(x) \in {g(X)}} \sum_{x \in {X}} p(x, g(x)) \log \frac{p(x, g(x))}{p(x) p(g(x))} \\ 
        &=H(Y)-H(Y \mid X)
\end{aligned}
\end{equation}

\subsubsection{Multi-view Mutual Information Maximization For Feature Selection }
Accessing a greater number of features can simplify prediction tasks, yet it necessitates larger models and increased parameterization, complicating dynamic feature selection.
This dilemma has been addressed under both non-uniform feature costs\cite{kachuee2018opportunistic} and fixed costs\cite{covert2023learning}, with our approach aligning with the latter. 
We capture a coefficient cube $ \hat{X} \in \mathcal{R}^{64 \times H/8 \times W/8} $ in the frequency domain, defaulting to 64 feature maps.
Now we have two variables, one is $g(X_I)$ for latent representation features while $I$ would be the batch size in practical, and another one is $\hat{X}_{I}^{J}$ while $J$ denotes the number of views that generated which is default in 64.
Utilizing an MI calculator, we estimate the MI between the latent feature $g(x)$ and each of the 64 views, re-ranking features based on MI scores for selection and loss computation.
Then the MI calculator could be defined as $I(g(X_I);\hat{X}_{I}^{J})$. 
Given the computational complexity of MI, we leverage MINE\cite{belghazi2018mine} as the estimator.
In line with frequency domain pruning literature\cite{xu2020learning}, \cite{chen2016compressing}, \cite{liu2018frequency}, we select the top $\sigma$ of DCT views, setting $\sigma$ as our fixed cost hyperparameter, for MI loss computation, as outlined in Equation \ref{eq:mi_loss}.

\begin{equation}
\begin{aligned}    
\label{eq:mi_loss}
    L_{MI}(g(X), \hat{X}) &= \underset{p(g(X)),p(\hat{X})}{\mathbb{E}} \left [ I(g(X);\hat{X})\right ] \\
    &= \frac{1}{B*J*\sigma\%} \sum_{i=0}^{B} \sum_{j=0}^{J*\sigma\%} I(g(X_i); \hat{X}_{i}^{j})
\end{aligned}
\end{equation}

Practically, our approach dynamically selects the most informative views for contrastive learning within a fixed-cost framework. 
Even though the cost remains constant, the feature selection adapts dynamically across epochs, ensuring each iteration of contrastive learning leverages the most correlated features.
The specifics of this dynamic selection process are demonstrated in the pseudocode outlined in Algorithm \ref{alg:mi}.


\begin{algorithm}[h]
\SetAlgoLined
\caption{MI-based Dynamic Multi-views Selection Pseudocode.}
    \KwIn{Encoded original feature map $g(X)$, 
                    Corresponded DCT coefficient feature map $\hat{X}$, 
                    MI estimator $MI(;)$, Cost hyperparameter $\sigma$} 
    \KwOut{MI loss $L_{MI}(g(X), \hat{X})$}
    initialize $b, v, h, w = \hat{X}.size()$, $loss\gets default list$ \;
    \For{$i = 1$ to $b$}
    {
        $loss(I)\gets default list$ \;
        \For{$j = 1~\textbf{to}~v$}
        {
            $loss(I).append(MI(g(X_i);\hat{X}_{i}^{j}))$
    `   }
        $loss(i).sort()$ \; 
        $loss(i) = mean(loss(i)(:v*\sigma\%))$ \;
    }
    $L_{MI}(g(X), \hat{X}) = mean(loss)$
\label{alg:mi}
\end{algorithm}

\subsection{Multi-view Contrastive Learning}

\subsubsection{Single-view Contrastive Loss}
The core objective of contrastive learning is to distinguish between positive (useful) and negative (useless) representations in an embedding space.
Traditional contrastive learning approaches\cite{gutmann2010noise}, \cite{mnih2013learning} typically select negative samples directly from the dataset, a method referred to as single-view contrastive learning.
In such a framework, given a dataset ${X|X=(x_1, x_2, ..., x_I)}$, one data point serves as the positive sample $x$. 
The remainder forms a contrastive memory bank $\hat{Y}$, culminating in a training batch $S={x, \hat{y}_{1}, \hat{y}2, ... ,\hat{y}{I-1}}$.
Within this set, we employ a discriminating function to evaluate the likelihood of a sample being positive. The associated contrastive loss is then formulated as in Equation \ref{eq:single_contrast}.

\begin{equation}    
\label{eq:single_contrast}
    L_{contrast}(g(X)) = \underset{S}{\mathbb{E}} \left [ log\frac{d(x)}{d(x) + {\textstyle \sum_{i=1}^{I-1} d(y_i)}}  \right ] 
\end{equation}

\subsubsection{Multi-view Contrastive Loss}
In multi-view contrastive learning, particularly in a two-view setting, the discriminating function is typically represented by a distance metric.
As demonstrated by Atito et al.\cite{atito2021sit}, the process commences by encoding representations into one-dimensional arrays via a compact neural network, in our case, consisting of three linear layers.
The similarity between these encoded representations is then quantified using a distance function, such as $\textit{Cosine Similarity}$, parameterized by a temperature hyperparameter $\tau$. 
Alternative metrics like $\textit{Euclidean Distance}$ and $\textit{Manhattan Distance}$ are also viable options.
In a scenario where the dataset offers two distinct views, the original set $v^1=X$ and an alternative view represented by DCT coefficient feature maps under a specific frequency order, $v^2=\hat{X}^1={\hat{x}^1_1, \hat{x}^1_2, ..., \hat{x}^1_I }$, the two-view contrastive loss is computed by summing the individual contrastive losses, with each view serving as the anchor in turn, as articulated in Equation \ref{eq:two_contrast}.

\begin{equation}   
\begin{aligned}
\label{eq:two_contrast}
    L_{contrast}(v^1,v^2) &= \underset{(v^1, v^2_1, v^2_2, ..., v^2_I)}{\mathbb{E}} \left [ log\frac{d(v^1,v^2)}{{\textstyle \sum_{i=1}^{I} d(v^1,v^2)}}  \right ] \\
    &+ \underset{(v^2, v^1_1, v^1_2, ..., v^1_I)}{\mathbb{E}} \left [ log\frac{d(v^2,v^1)}{{\textstyle \sum_{i=1}^{I} d(v^2,v^1)}}  \right ] \\
    s.t. ~~d(v^1,v^2) &= exp( \frac{v^1 \cdot v^2}{\left \| v^1 \right \| \cdot \left \| v^2 \right \| } \cdot \frac{1}{\tau})
\end{aligned}
\end{equation}

In our multi-view contrastive learning framework, particularly under the self-supervised learning paradigm, we default to utilizing 64 views.
For each epoch, the model processes the original input set $X$ alongside its corresponding DCT view set $\hat{X}$.
The contrastive loss, optimized in this context, is formulated as in Equation \ref{eq:contrast_loss_self}.
This formulation considers a batch size of $B$ and incorporates only the selected number of views, $M=64*\sigma\%$, ensuring that the learning process is both focused and computationally efficient.

\begin{equation}    
\label{eq:contrast_loss_self}
    L_{contrast}(g(X), \hat{X}) = \sum_{\underset{j\ne k}{1<(j,k)<B} } \textstyle \sum_{i=0}^{M} L_{contrast}(X_j,\hat{X}^i_j)
\end{equation}

It is also easy to transition to a semi-supervised learning paradigm, the framework would take advantage of the ground truth information, denoted as the set $Y$.
Within this context, the contrastive loss is recalibrated to incorporate this additional insight, leading to an optimized formulation as presented in Equation \ref{eq:contrast_loss_semi}.
This integration of ground truth data into the learning process aims to further refine and enhance the model's performance.

\begin{equation}    
\begin{aligned}
\label{eq:contrast_loss_semi}
    L_{contrast}(g(X), \hat{X}) &= \sum_{\underset{j\ne k}{1<(j,k)<B}} \textstyle \sum_{i=0}^{M} L_{contrast}(X_j,\hat{X}^i_j) \\ 
                    &+ \sum_{\underset{j\ne k}{1<(j,k)<B}} L_{contrast}(X_j,Y_k)    
\end{aligned}
\end{equation}

The pseudocode for this partition is demonstrated as Algorithm \ref{alg:contrast}.

\begin{algorithm}[h]
\SetAlgoLined
\caption{Selected Multi-views Contrastive Learning Pseudocode.}
\KwIn{Encoded original feature map $g(X)$, 
                Selected multi-views in frequency domain $\hat{X}$, 
                Distance estimator $d(,)$, Contrastive loss calculator $CL(\cdot)$,
                Temperature hyperparameter $\tau$.}
\KwOut{Contrastive loss $L_{CL}(g(X), \hat{X})$.}
initialize $b, v, h, w = \hat{X}.size()$, $loss\gets default list$\\ 
\For{$i = 1~\textbf{to}~b$}
{
    $loss(i)\gets default list$ \\
    \For{$j = 1~\textbf{to}~v$}
    {
        $l_1 = CL(d(g(X_i), \hat{X}_i^{j}))$\\
        \uIf {\textit{Self-supervise}}
        {
            $loss(i).append(l_1)$
        } 
        \uElseIf {\textit{Semi-supervise}}
        {
            $l_2 = CL(d(X_i, Y_i))$\\
            $loss(i).append(l_1 + l_2)$
        }
    }
}
$L_{CL}(g(X), \hat{X}) = mean(loss)$
\label{alg:contrast}
\end{algorithm}

\subsection{Segmentation Framework}
\begin{figure*}[ht]
\centering
\includegraphics[width=10cm]{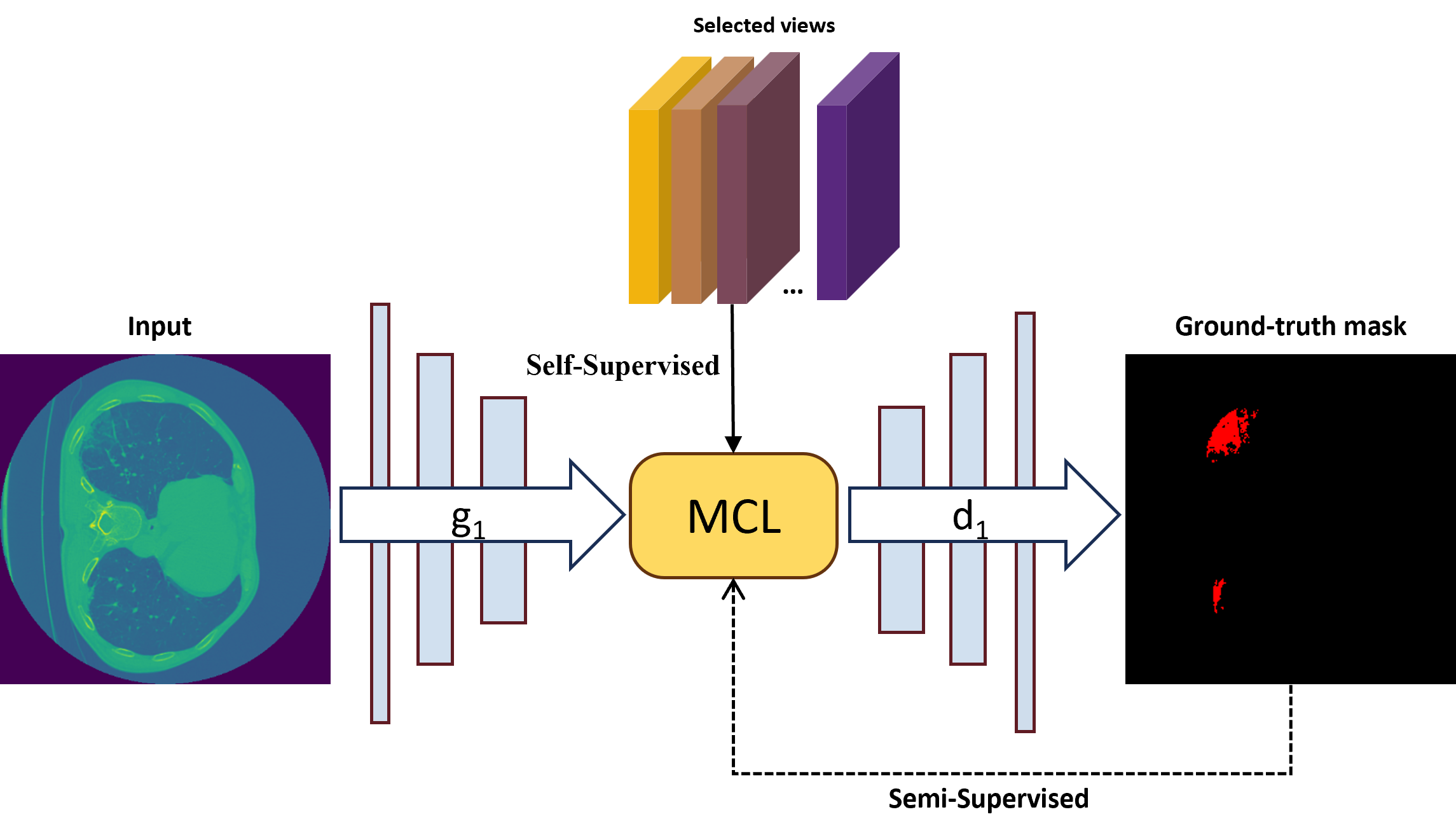}
\caption{The proposed framework introduces a multi-view contrastive learning strategy executed within the latent space, utilizing selectively chosen views to refine representation learning. Although Cosine similarity is the primary distance metric in our setting, the framework's design permits the adoption of alternative metrics, such as the L1 norm, L2 norm, or others, depending on the specific requirements. Given a generated multi-view DCT cube, the selection is based on its proximity to the original image representation within the information entropy space.} 
\label{fig:framework}
\end{figure*}

The architecture of our comprehensive framework is depicted in Fig.\ref{fig:framework}.
Retaining the original backbone, our approach primarily focuses on maximizing MI for multi-view selection toward more effective contrastive learning, effectively distilling useful representations.
Experimentally, the embeddable of our framework is demonstrated through its application to purely CNN-based blocks, leveraging a U-Net\cite{ronneberger2015u} architecture, as well as to an additional Transformer block, embodied by the TranUNet\cite{chen2021transunet} architecture.

Upon generating DCT coefficient cubes through an offline transformation, they are inputted alongside original CT images into our framework.
The original images are encoded via three CNN-block layers, or an additional ViT embedding layer followed by three CNN-block layers when integrating with the TransUNet architecture.
This results in a representation of the shape $g(X)\in \mathcal{R}^{b\times c\times \frac{h}{8} \times \frac{w}{8}}$.
A parallel structure, employing three pure-CNN blocks but with distinct weights, downsamples the DCT coefficients to a comparable representation.
Subsequently, our Mutual Information Maximization-based Multi-view Contrastive learning (MIMIC) strategy bridges these two representations, fostering a more effective representation learning.
The refined representation is then fed into a conventional decoder to generate the predicted lesion masks. 
The segmentation performance is quantified by computing the $\textit{BCE Loss}$ and $\textit{DICE Loss}$ between the predicted and ground-truth masks, culminating in the total loss as defined in Equation \ref{eq:total_loss}.

\begin{equation}    
\begin{aligned}
\label{eq:total_loss}
    Loss_{total} &= L_{\textit{BCE}}(X,Y) + L_{\textit{DICE}}(X,Y) \\ 
    & + L_{contrast}(g(X), \hat{X}, Y) + L_{MI}(g(X),\hat{X})    
\end{aligned}
\end{equation}

It's worth highlighting that within our MIMIC module, while we aim to maximize MI between $g(X)$ and $\hat{X}$, we concurrently minimize the representation distance for contrastive learning. 
This dual objective may seem counterintuitive. However, insights from the variational bound of mutual information\cite{oord2018representation}, \cite{poole2019variational}, \cite{tian2020contrastive}, elucidate this problem.
When the distance function $d(,)$ is optimized, its value aligns proportionally with the density ratio between the joint distribution of the two entities $p(g(X), \hat{X})$ and the individual marginal distributions $p(g(X))$ and $p(\hat{X})$.
Referring to our Equation \ref{eq:two_contrast}, the expected value is essentially similar to point-wise mutual information.
This conceptual alignment is mathematically substantiated and gets the  Equation. \ref{eq:mi_bound}, reconciling between maximizing MI and minimizing representation distance within the MIMIC module.

\begin{equation}    
\begin{aligned}
\label{eq:mi_bound}
    d(g(X),\hat{X}) & \propto \frac{p(g(X),\hat{X})}{p(X)p(\hat{X})} \propto \frac{p(g(X)|\hat{X})}{p(X)} \\ 
    I(g(X);\hat{X}) & \ge \lg{(M)} - L_{contrast}(g(X), \hat{X})    
\end{aligned}
\end{equation}

$M$ denotes the size of the memory bank, representing the count of selected views within the self-supervised MI maximization setting.
Drawing from Eq. \ref{eq:mi_bound}, it becomes evident that minimizing the contrastive loss intrinsically maximizes the lower bound of the MI between $g(X)$ and $\hat{X}$. 
This interconnection aligns with the objectives of our contrastive learning approach, ensuring that the process of minimizing loss is tied with the enhancement of mutual information in the learned representations.

\section{Implement Details}
\subsection{Datasets}
This study utilizes three distinct datasets for COVID-19 CT segmentation: COVID19-CT-100\cite{medseg2021medseg}, COVID19-CT-Seg20\cite{jun2020covid}, and MosMedData\cite{morozov2020mosmeddata}.
The COVID19-CT-100 dataset, collected by the Italian Society of Medical and Interventional Radiology, comprises 100 CT slices from over 40 patients, all confirmed to have COVID-19.
This dataset is particularly notable for its detailed annotations of various infection regions, such as ground-glass opacity (GGO) and consolidation.
The COVID19-CT-Seg20 dataset encompasses 20 CT images with annotations by two radiologists and validation by a third, experienced radiologist, focusing on lung and infection areas.
Our research concentrates specifically on infection segmentation, given its complexity and clinical relevance.
Additionally, the MosMedData, assembled by Moscow's Research and Practical Clinical Center for Diagnostics and Telemedicine Technologies, contains 50 CT scans with less than 25\% lung infections, meticulously labeled by experts.
Due to the limited number of scans and the substantial inter-slice spacing in these volumetric datasets, our study adopts a 2D segmentation approach, as per prior work\cite{jia2023convolutional}, so we just compare with the baseline and metrics for 2D slices under 2D pre-processing.
Consequently, we processed 100, 1844, and 785 2D CT slices from the COVID19-CT-100, COVID19-CT-Seg20, and MosMedData datasets, respectively.
It's important to note that this study does not examine the impact of subject variables such as age, gender, or race on the results, as this information was not disclosed by the dataset providers. 
The specifics of these datasets are systematically presented in Table \ref{tb:data}.

\begin{table}[t]
    \centering
    \caption{Detailed information about the three datasets.}
    \begin{tabular}{lcc} \hline
    Datasets & Slice number & Original resolution\\ \hline
    COVID19-CT-100\cite{medseg2021medseg}& 100 & $512 \times 512$ \\ 
    COVID19-CT-Seg20\cite{jun2020covid} & 1844 &  $512 \times 512-630 \times 630$ \\ 
    MosMedData\cite{morozov2020mosmeddata}  & 785 & $512 \times 512$ \\ \hline
    \end{tabular}
    \label{tb:data}
\end{table}

\subsection{Implementation Details}
During the training phase, we implemented on-the-fly data augmentation to mitigate potential overfitting. 
This included random scaling between 0.8 and 1.2, rotation within $(\pm15^{\circ})$, intensity shifts of $(\pm0.1)$, and scaling between 0.9 and 1.1. 
Post-augmentation, each image was resized to $256 \times 256$ pixels using bilinear interpolation.
The training process spanned 300 epochs, incorporating a linear warm-up during the initial 5 epochs and an early stopping mechanism activated if the total loss did not decrease over 50 consecutive epochs.
Model optimization was conducted using the Adam optimizer, synchronized batch normalization, and a batch size of 20. The learning rate was initiated at $5e^{-4}$ and followed a decay pattern of $(1 - \frac{current_epoch}{max_epoch})^{0.9}$.
Furthermore, $l_2$ weight decay regularization of $5e^{-5}$ was applied to enhance model generalization.
It's worth noting that experiments were conducted independently across the three datasets, data from different datasets were not merged for training purposes. 
All experimental procedures were executed using PyTorch 1.6.0 and deployed on a workstation equipped with 2 NVIDIA TITAN XP GPUs.

\begin{table*}[t]s
\centering
\caption{Quantitative results of different methods the dataset. The best and second best results are shown in \textcolor{red}{red} and \textcolor{blue}{blue}, respectively. The values of DSC and IoU are in percentage terms.}
\begin{tabular}{cccccccccccccc}
\hline
\multirow{2}*{\textbf{Methods}} & \multicolumn{4}{c}{COVID19-CT-100} & \multicolumn{4}{c}{COVID19-CT-Seg20} & \multicolumn{4}{c}{MosMedData}\\ 
\textbf{} & DSC$\uparrow$ & mIoU$\uparrow$ & HD95$\downarrow$ & ASD$\downarrow$ & DSC$\uparrow$ & mIoU$\uparrow$ & HD95$\downarrow$ & ASD$\downarrow$ & DSC$\uparrow$ & mIoU$\uparrow$ & HD95$\downarrow$ & ASD$\downarrow$\\ \hline

FCN-8s & 59.32 & 52.06 & 8.11 & 1.50 & 78.73 & 74.10 & 8.83 & 1.19 & 60.39 & 57.29 & 6.45 & 1.37 \\ 
Deeplabv3+ & 71.59 & 62.83 & 3.21 & 0.42 & 69.16 & 73.24 & 1.12 & 0.27 & 65.23 & 64.88 & 7.97 & 1.04 \\ \hline

U-Net & 72.13 & 63.78 & 2.45 & 0.37 & 78.47 & 73.48 & 1.04 & \color{blue}0.16 & 69.20 & 67.84 & 4.55 & 0.93 \\ 
UNet++ & 68.75 & 60.77 & 2.83 & 0.40 & 78.66 & 73.78 & 1.02 & 0.25 & 68.32 & 65.49 & 6.43 & 1.00 \\ 
ResUNet & 70.31 & 61.40 & 3.00 & \color{blue}0.33 & 77.95 & 72.60 & 1.08 & 0.23 & 66.58 & 63.30 & 5.42 & 0.97 \\ 
AttUNet & 74.35 & 65.77 & 2.27 & 0.51 & 77.25 & 71.66 & 1.82 & 0.34 & 68.70 & 65.80 & \color{blue}5.41 & 1.14 \\ 
Inf-Net & 76.44 & 71.41 & 2.89 & 0.35 & 72.33 & 65.16 & 1.76 & 0.23 & 63.48 & 65.23 & 7.33 & 1.24 \\
MiniSeg & \color{blue}76.91 & \color{blue}73.26 & 2.30 & 0.42 & 78.27 & 76.22 & \color{blue}1.00 & 0.28 & 72.87 & 70.28 & 5.45 & 1.08 \\ \hline

TransUNet & 65.45 & 54.64 & \color{blue}0.68 & 0.39 & 77.43 & 72.83 & 1.14 & 0.22 & 67.19 & 61.19 & 7.05 & 1.00 \\ 
MedT & 68.05 & 59.92 & 4.00 & 0.70 & 75.93 & 69.58 & 1.28 & 0.17 & 66.72 & 58.85 & 6.49 & 1.11 \\ 
MISSFormer & 71.35 & 62.45 & 2.24 & 0.42 & 77.59 & 72.65 & 1.04 & \color{blue}0.16 & 65.52 & 54.47 & 6.68 & 1.17 \\ \hline

UNet+MIMIC & \color{red}82.20 & \color{red}77.28 & 2.24& \color{red}0.32 & \color{blue}84.08 & \color{red}80.76 & \color{red}0.96 & \color{red}0.14 & \color{red}80.32 & \color{red}78.16 & \color{red}3.32 & \color{blue}0.92\\ 
TransUNet+MIMIC & 74.38 & 66.11 & \color{red}0.62& \color{blue}0.33 & \color{red}88.26& \color{blue}78.25 & 1.02 & \color{blue}0.16 & \color{blue}77.43 & \color{blue}76.78 & \color{blue}5.41& \color{red}0.86\\ \hline
\end{tabular}
\label{tb:main}
\end{table*}

\begin{figure*}[htp]
\centering
\includegraphics[width=16cm]{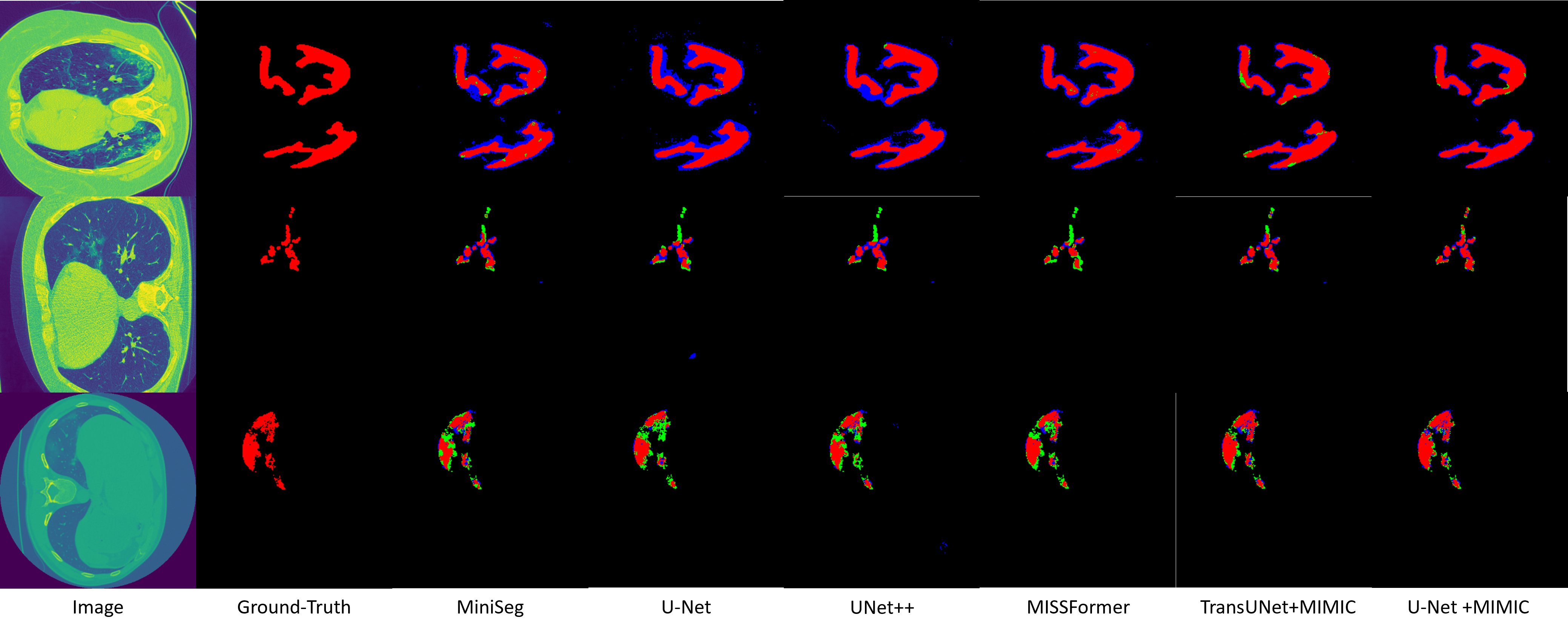}
\caption{Visualization of the segmentation results on COVID19-CT-Seg20 (row 1), COVID19-CT-100(row 2), and MosMedData (row 3) three datasets produced by our proposed MIMIC framework under two backbones and other 4 competitive baselines. The true positive, false negative and false positive are highlighted with \textcolor{red}{red}, \textcolor{green}{green}, and \textcolor{blue}{blue}, respectively.} 
\label{fig:main_vis}
\end{figure*}

\subsection{Baseline and Evaluation Metrics}
Our approach was benchmarked against eleven segmentation models to validate its effectiveness.
These included FCN-8s\cite{long2015fully}, DeepLabv3+\cite{chen2018encoder}, U-Net++\cite{zhou2019unet++}, Att-UNet\cite{oktay2022attention}, Res-UNet\cite{xiao2018weighted}, the original U-Net\cite{ronneberger2015u}, MiniSeg\cite{qiu2021miniseg}, Inf-Net\cite{fan2020inf}, TransUNet\cite{chen2021transunet}, MedT\cite{valanarasu2021medical}, and MISSFormer\cite{huang2022missformer}.
The original U-Net and TransUNet were recognized for their robust performance in 2D medical image segmentation.
U-Net++, Att-UNet, and Res-UNet have been influential U-Net-based models in this domain, while MedT and MISSFormer are renowned for their transformer-based architectures. 
FCN-8s and DeepLabv3+ are widely acknowledged in general image semantic segmentation, with MiniSeg and Inf-Net emerging as state-of-the-art models in COVID-19 segmentation.
Performance evaluation was conducted using four metrics: mean intersection over union (mIoU), Dice similarity coefficient (DSC), Average surface distance (ASD), and $5\%$ Hausdorff distance (HD95).
mIoU and DSC, as overlap-based metrics, are expressed as percentages, with higher values signifying superior performance.
In contrast, ASD and HD95, as shape distance-based metrics, measure the dissimilarity between the segmentation results and the ground truth, where lower values denote more accurate segmentation.

\section{Experiment Results}
\begin{table*}[t]
\centering
\caption{Ablation study of different methods the dataset. The best results are shown in \textcolor{red}{red}. The values of DSC and IoU are in percentage terms.}
\begin{tabular}{lccccccccccccc}
\hline
\multirow{2}*{\textbf{Methods}} & \multicolumn{4}{c}{COVID19-CT-100} & \multicolumn{4}{c}{COVID19-CT-Seg20} & \multicolumn{4}{c}{MosMedData}\\ 
\textbf{} & DSC$\uparrow$ & mIoU$\uparrow$ & HD95$\downarrow$ & ASD$\downarrow$ & DSC$\uparrow$ & mIoU$\uparrow$ & HD95$\downarrow$ & ASD$\downarrow$ & DSC$\uparrow$ & mIoU$\uparrow$ & HD95$\downarrow$ & ASD$\downarrow$\\ \hline

U-Net & 72.13 & 63.78 & 2.45 & 0.37 & 78.47 & 73.48 & 1.04 & 0.16 & 69.20 & 67.84 & 4.55 & 0.93 \\ 
U-Net+MI & 76.63 & 65.80 & 2.31 & 0.64 & 80.39 & 77.88 & 1.02 & 0.17 & 66.35 & 60.27 & 7.51 & 1.00 \\ 
U-Net+CL & 77.29 & 68.55 & 2.33 & 0.38 & 80.62 & 75.30 & 0.97 & 0.14 & 78.06 & 75.29 & 4.69 & 0.92 \\ 
U-Net+MIMIC & \color{red}82.20 & \color{red}77.28 & \color{red}2.24& \color{red}0.32 & \color{red}84.08 & \color{red}80.76 & \color{red}0.96 & \color{red}0.14 & \color{red}80.32 & \color{red}78.16 & \color{red}3.32 & \color{red}0.92 \\ \hline

TransUNet & 65.45 & 54.64 & 0.68 & 0.39 & 77.43 & 72.83 & 1.14 & 0.22 & 67.19 & 61.19 & 7.05 & 1.00 \\ 
TransUNet+MI & 69.23 & 60.31 & 0.66 & 0.34 & 86.20 & 77.92 & 1.12 & 0.20 & 70.22 & 66.55 & 7.01 & 0.99 \\ 
TransUNet+CL & 70.17 & 62.07 & 0.66 & 0.36 & 85.38 & 76.48 &  1.05 & 0.17 & 73.23 & 69.35 & 6.30 & 0.91 \\ 
TransUNet+MIMIC & \color{red}74.38 & \color{red}66.11 & \color{red}0.62& \color{red}0.33 & \color{red}88.26& \color{red}78.25 & \color{red}1.02 & \color{red}0.16 & \color{red}77.43 & \color{red}76.78 & \color{red}5.41& \color{red}0.86 \\ \hline
\end{tabular}
\label{tb:ablation}
\end{table*}

\begin{figure*}[htp]
\centering
\includegraphics[width=14cm]{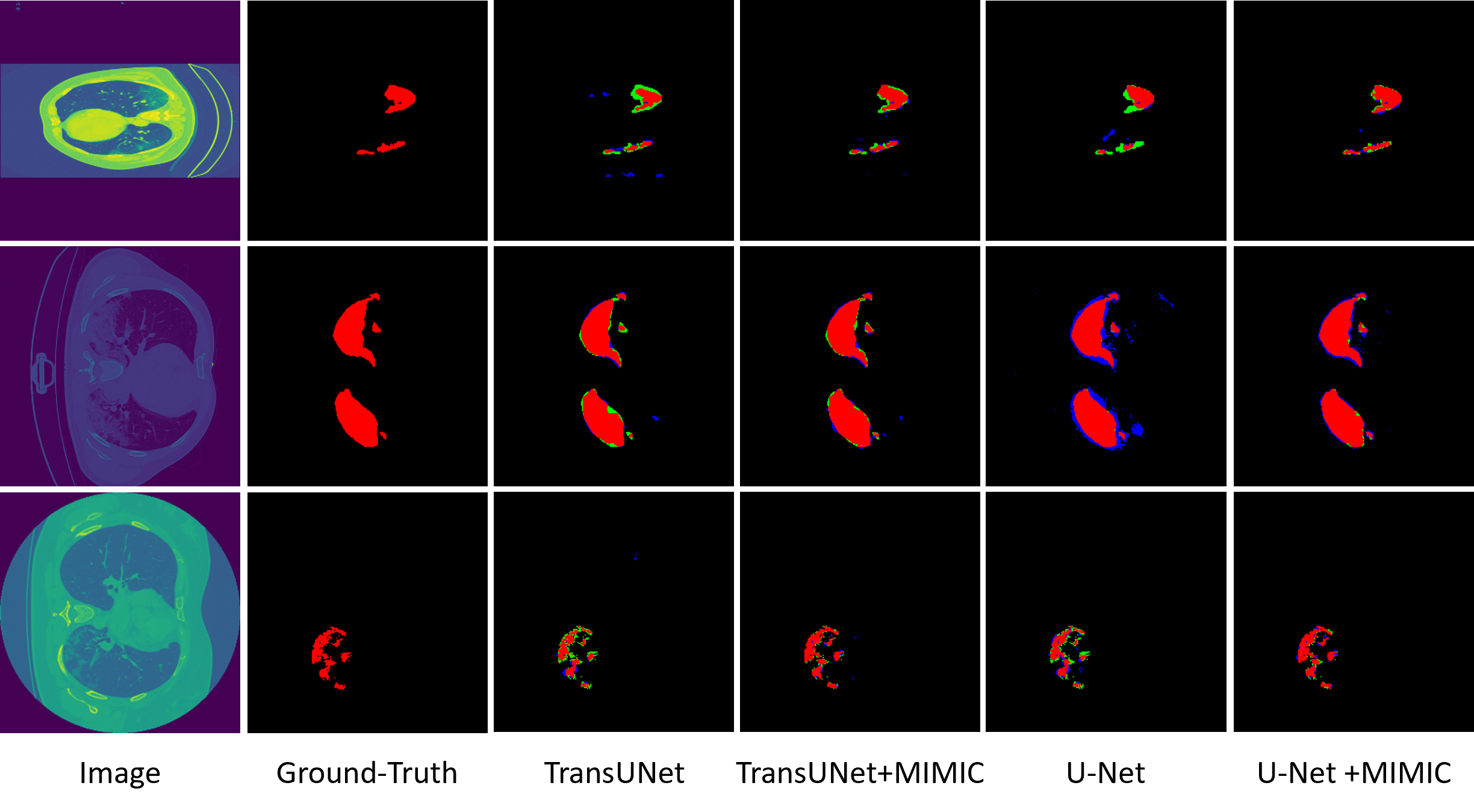}
\caption{Visualization of the segmentation results in ablation study on COVID19-CT-100 (row 1), COVID19-CT-Seg20 (row 2), and MosMedData (row 3) three datasets produced by our proposed MIMIC framework under two backbones and their used baselines. The true positive, false negative, and false positive are highlighted with \textcolor{red}{red}, \textcolor{green}{green}, and \textcolor{blue}{blue}, respectively.} 
\label{fig:ab_vis}
\end{figure*}

\subsection{Comparative Experiments}
In this section, our MIMIC framework was benchmarked against other state-of-the-art (SOTA) and influential segmentation networks, with the findings detailed in Table \ref{tb:main}.
Notably, our method outperformed all baselines when configured with identical backbones.
Specifically, U-Net+MIMIC and TransUNet+MIMIC exhibited superior performance across all metrics.
It's crucial to highlight that, contrary to common trends in Computer Vision tasks where Transformer-based methods usually excel, our experiments indicated a slight underperformance. 
We attribute this primarily to dataset size constraints. 
Transformer models, typically characterized by larger parameter sets, demand extensive and diverse datasets to fully harness their potential.
This is particularly challenging in medical imaging, where data often exhibits feature sparsity and clinical manifestations present long-tail distribution challenges\cite{zhou2021review}.
Conventionally, Vision Transformer models\cite{han2022survey} are trained on vast datasets\cite{deng2009imagenet, lin2014microsoft, geiger2012we}, and three Transformer-based baselines\cite{chen2021transunet, huang2022missformer, valanarasu2021medical} are also trained on collections nearing 3000 samples.
Contrastingly, the largest dataset in our study, COVID19-CT-Seg20, comprises only 1844 slices.
However, our novel multi-view data augmentation approach, combined with a contrastive learning pipeline, effectively addresses these challenges.
It optimally utilizes limited data to derive more meaningful representations, significantly enhancing the efficacy of the TransUNet backbone when integrated with our MIMIC module.
Moreover, on the smaller dataset COVID19-CT-100, Transformer-based methods demonstrated improved performance on distance-related metrics, resonating with their inherent proficiency in capturing long-range relationships.

\begin{figure*}[h]  
\begin{center}  
    \subfigure[DSC and hyperparameter $\delta$]
    {   
        \includegraphics[width=0.23\linewidth]{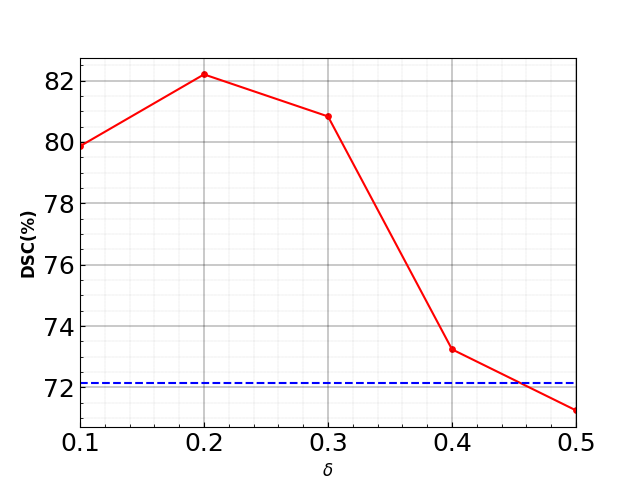}  
    }  
    \subfigure[mIoU and hyperparameter $\delta$]
    {
        \includegraphics[width=0.23\linewidth]{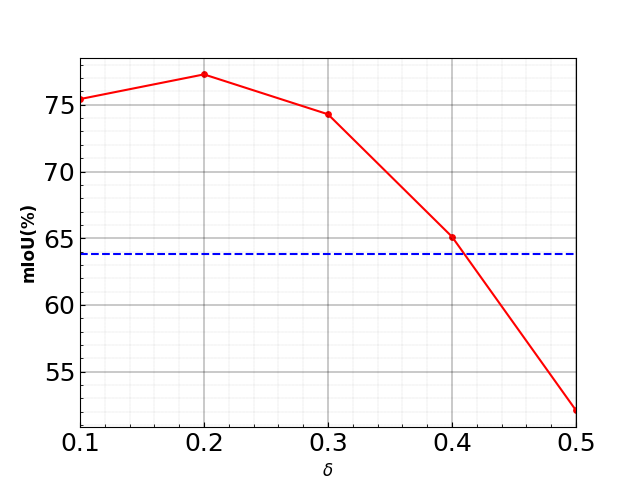}  
    }  
    \subfigure[HD95 and hyperparameter $\delta$]
    {
        \includegraphics[width=0.23\linewidth]{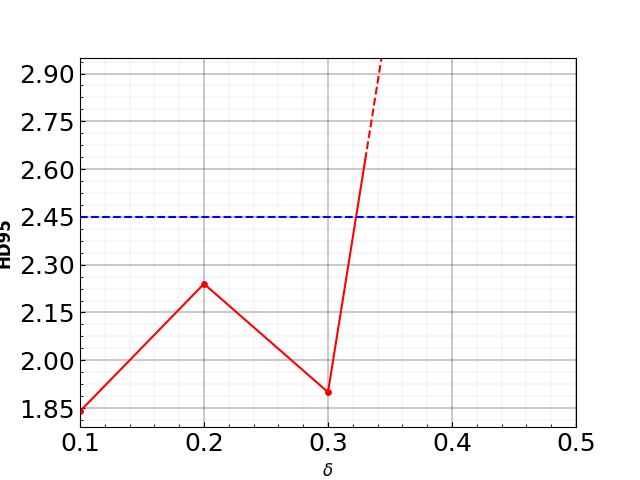}  
    } 
    \subfigure[ASD and hyperparameter $\delta$]
    {
        \includegraphics[width=0.23\linewidth]{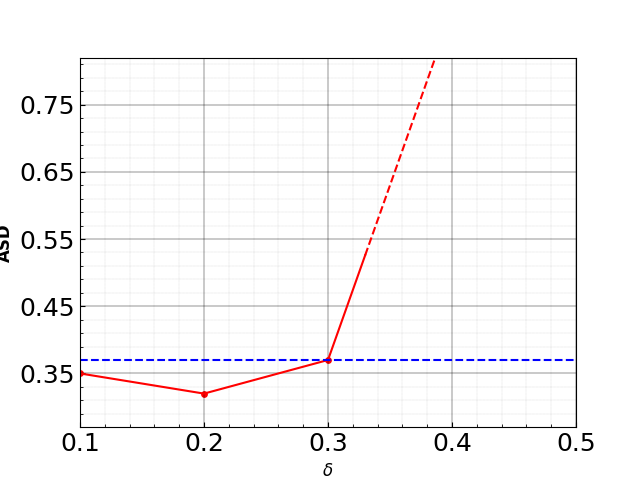}  
    } 
    \caption{Impact of $\delta$ on segmentation performance when using the U-Net as the backbone, the \textcolor{blue}{dashlines} are the baseline results where the proposed MIMIC module is not integrated. Higher DSC and mIoU values or lower HD95 and ASD values indicate better performance.}  
    \label{para:u_d}
\end{center} 
\end{figure*}

\begin{figure*}[h]  
\begin{center}  
    \subfigure[DSC and hyperparameter $\delta$]
    {   
        \includegraphics[width=0.23\linewidth]{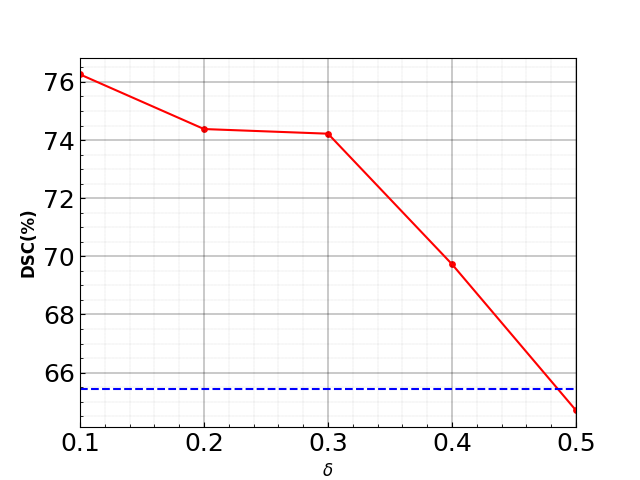}  
    }  
    \subfigure[mIoU and hyperparameter $\delta$]
    {
        \includegraphics[width=0.23\linewidth]{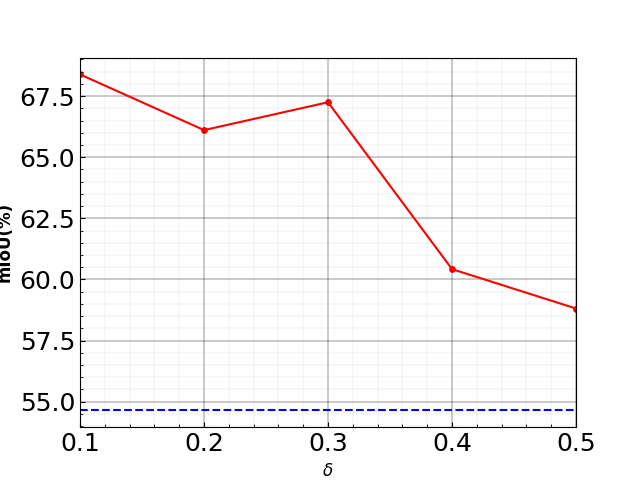}  
    }  
    \subfigure[HD95 and hyperparameter $\delta$]
    {
        \includegraphics[width=0.23\linewidth]{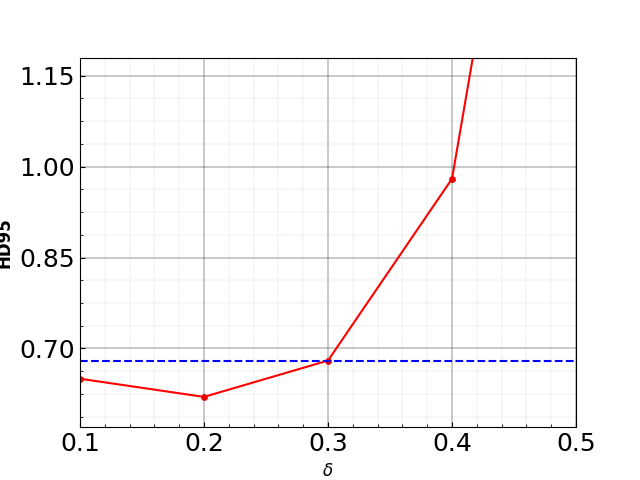}  
    } 
    \subfigure[ASD and hyperparameter $\delta$]
    {
        \includegraphics[width=0.23\linewidth]{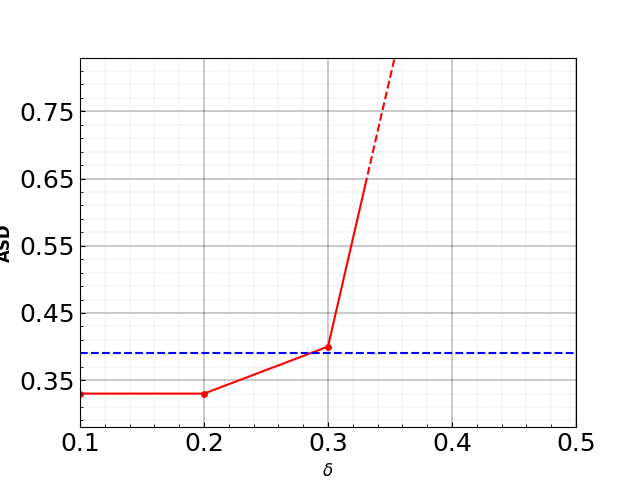}  
    } 
    \caption{Impact of $\delta$ on segmentation performance when using the TransUNet as the backbone, the \textcolor{blue}{dashlines} are the baseline results where the proposed MIMIC module is not integrated. Higher DSC and mIoU values or lower HD95 and ASD values indicate better performance.}  
    \label{para:t_d}
\end{center} 
\end{figure*}

\subsection{Ablation Study}
An ablation study was executed across all three datasets to ascertain the effectiveness of our MIMIC module. 
Our model was divided into two primary components for this study: the first solely leveraged the Mutual Information (MI) maximization strategy, where mutual information loss was calculated based on the top $\delta\%$ of features, and the second did not optimize the mutual information loss yet persisting in feature selection and contrastive learning. 
In this context, the hyperparameters were standardized, with $\delta$ set at 5 and the default patch size at 8.
The impact of these distinct strategies was examined using two backbone architectures, U-Net and TransUNet. 
Consequently, U-Net+MIMIC and TransUNet+MIMIC were compared against their baseline and the aforementioned two separated scenarios individually, with the comparative results detailed in Table \ref{tb:ablation}.
This suggests that while augmenting multi-view features, a discernment of positive from negative features is critical.
Even contrastive learning emerged as a promising approach, it was evident that learning from more meaningful samples after MI strategy is better. 
This illustrates the significance of our MIMIC module, which synergistically combines these two elements, fostering a more robust and effective representation learning method.

\subsection{Prediction Visualization}
To further illustrate the effectiveness of our MIMIC module, alongside the statistical results above, we present a series of visualization pictures.
Fig. \ref{fig:main_vis} showcases a case study derived from three representative samples across the three datasets, featuring input images, ground truth, and the segmentation masks predicted by various methods.
Additionally, Fig. \ref{fig:ab_vis} offers a focused visualization of the ablation study, comparing the results of U-Net+MIMIC and TransUNet+MIMIC against their respective baselines.
These visual comparisons demonstrate that our MIMIC module not only enhances the network's ability to detect precise edges but also improves to capture spatial semantic information.
For instance, while the original U-Net tends to classify background or healthy lung areas as lesions, and TransUNet overlooks some lesions, integrating our MIMIC module significantly resolves these issues, yielding more accurate and reliable segmentation.

\subsection{Other Experiments}
\begin{figure*}[h]  
\begin{center}  
    \subfigure[DSC and hyperparameter $\textit{p}$]
    {   
        \includegraphics[width=0.23\linewidth]{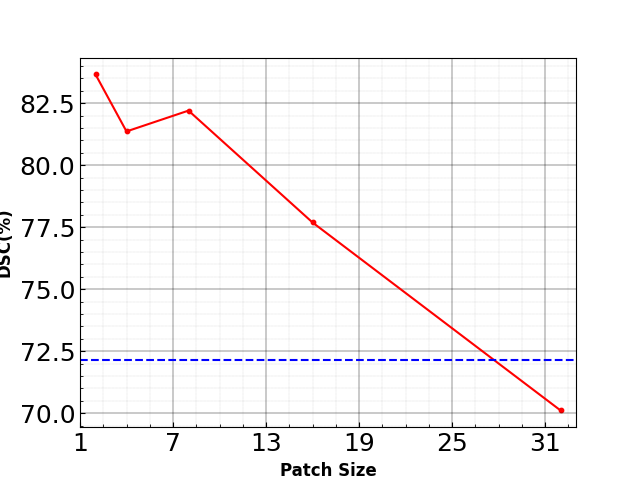}  
    }  
    \subfigure[mIoU and hyperparameter $\textit{p}$]
    {
        \includegraphics[width=0.23\linewidth]{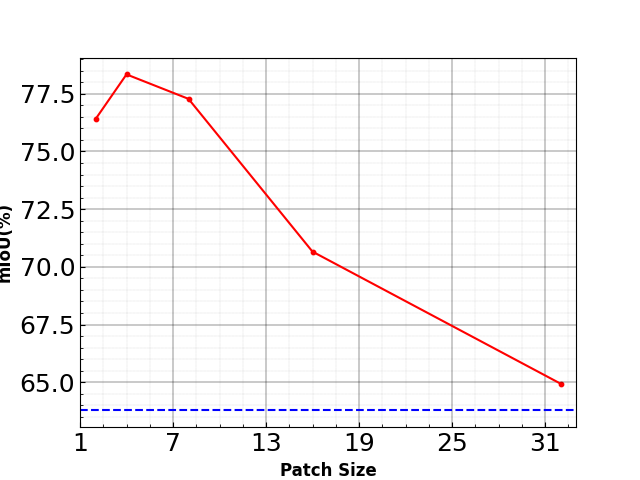}  
    }  
    \subfigure[HD95 and hyperparameter $\textit{p}$]
    {
        \includegraphics[width=0.23\linewidth]{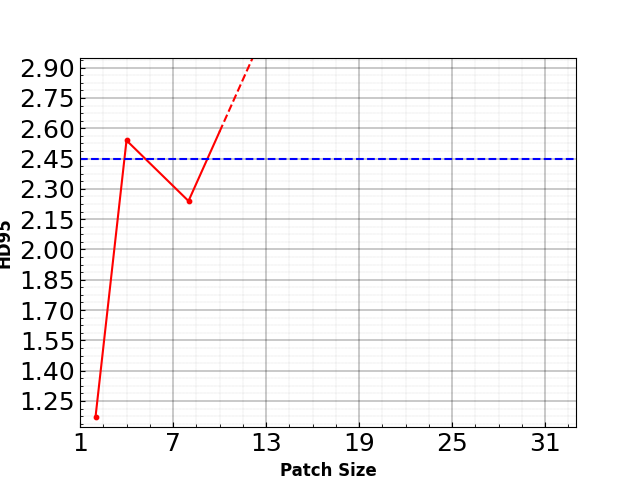}  
    } 
    \subfigure[ASD and hyperparameter $\textit{p}$]
    {
        \includegraphics[width=0.23\linewidth]{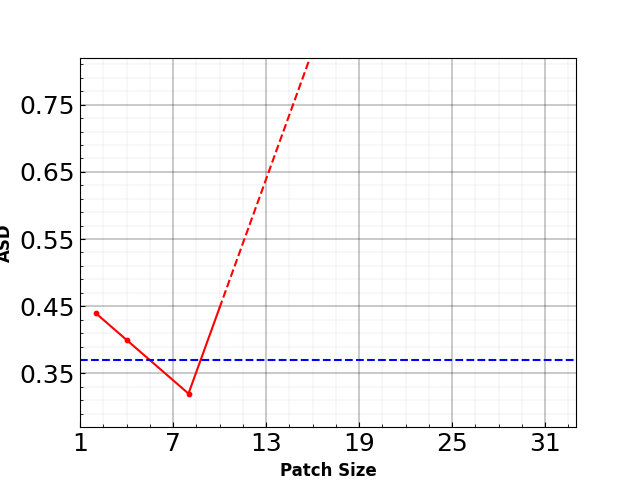}  
    } 
    \caption{Impact of $\textit{p}$ on segmentation performance when using the U-Net as the backbone, the \textcolor{blue}{dashlins} are the baseline results where the proposed MIMIC module is not integrated. Higher DSC and mIoU values or lower HD95 and ASD values indicate better performance.}  
    \label{para:u_p}
\end{center} 
\end{figure*}

\begin{figure*}[h]  
\begin{center}  
    \subfigure[DSC and hyperparameter $\textit{p}$]
    {   
        \includegraphics[width=0.23\linewidth]{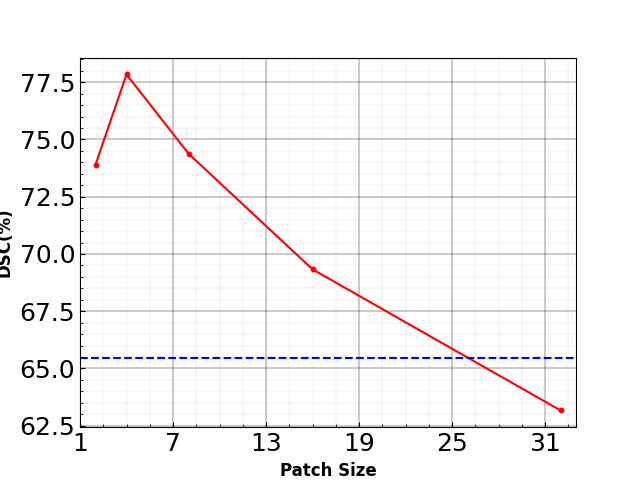}  
    }  
    \subfigure[mIoU and hyperparameter $\textit{p}$]
    {
        \includegraphics[width=0.23\linewidth]{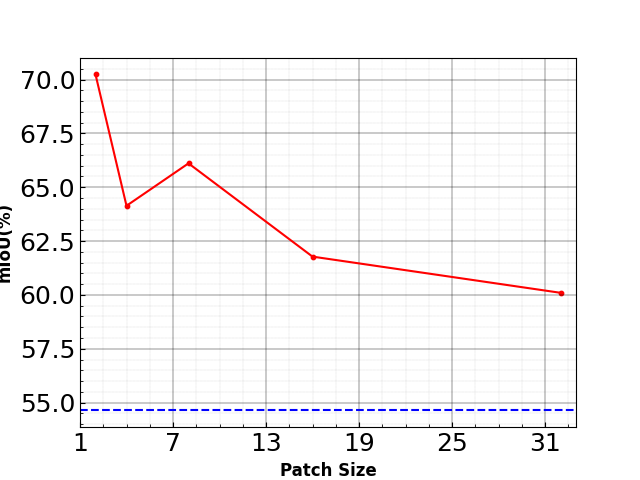}  
    }  
    \subfigure[HD95 and hyperparameter $\textit{p}$]
    {
        \includegraphics[width=0.23\linewidth]{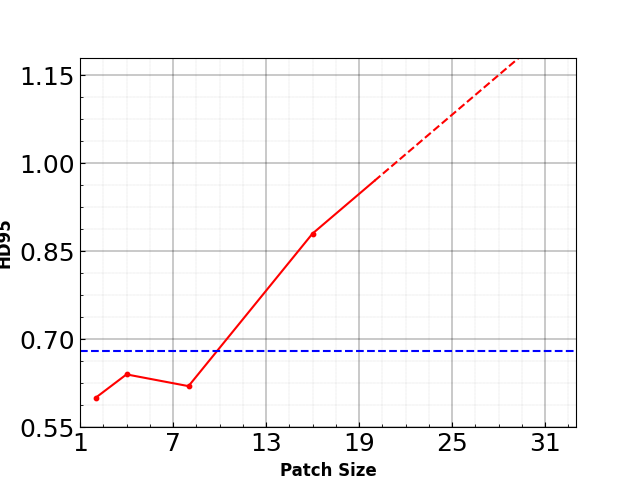}  
    } 
    \subfigure[ASD and hyperparameter $\textit{p}$]
    {
        \includegraphics[width=0.23\linewidth]{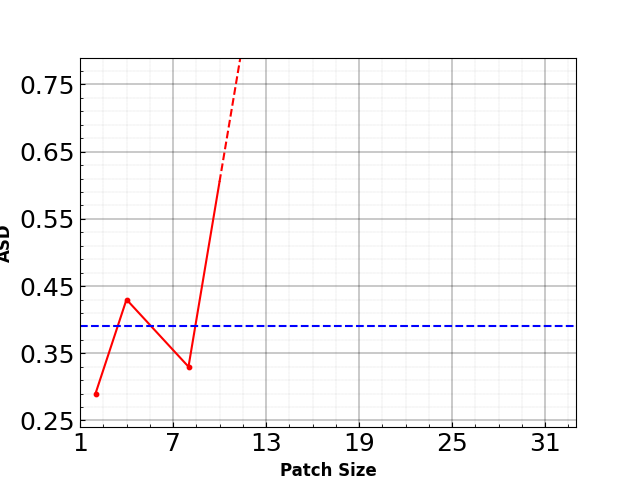}  
    } 
    \caption{Impact of $\textit{p}$ on segmentation performance when using the TransUNet as the backbone, the \textcolor{blue}{dashlines} are the baseline results where the proposed MIMIC module is not integrated. Higher DSC and mIoU values or lower HD95 and ASD values indicate better performance.}  
    \label{para:t_p}
\end{center} 
\end{figure*}

An in-depth analysis was performed to evaluate the influence of two principal hyperparameters, $\textit{p}$ and $\delta$, on segmentation performance, with the findings illustrated in Fig. \ref{para:u_d}-\ref{para:t_p}. 
During the experiments, $\textit{p}$ was fixed at 8, while $\delta$ varied within the range [0.1, 0.5] with steps of 0.1. For $\textit{p}$, the values were chosen as integer powers of 2, spanning from 2 to 32, to facilitate straightforward slicing of the original image.
The results generally outperformed the baselines across various parameter settings.
Meanwhile, the findings are similar to some related studies\cite{liu2018frequency, xu2020learning}, it was observed that the majority of frequency domain representations were less meaningful.
Consequently, setting $\delta$ below 0.3 yielded significantly improved results by minimizing the inclusion of redundant representations in the contrastive learning process.
Similarly, a smaller $\textit{p}$ enabled the extraction of more fine-grained representations, thereby enhancing performance.
In summary, the most robust and optimal outcomes across all four metrics were achieved when $\textit{p}$ and $\delta$ were set to 8 and 0.2, respectively, highlighting the critical role of select appropriate hyperparameters in refining segmentation performance.

\section{Conclusion}
In this study, we introduce a novel self-supervised contrastive learning approach, augmented by a mutual information-based feature selection mechanism, for enhanced representation learning in medical image segmentation.
Our proposed module seamlessly integrates with existing models, improving segmentation performance under four different metrics.
The method capitalizes on multi-view feature extraction within the frequency domain.
A continuous mutual information maximization strategy selects the most useful and informative representations according to the information entropy space.
These chosen representations are then employed in a self-supervised contrastive learning process, effectively refining the learning of representations for subsequent segmentation tasks.
Our methodology is estimated under rigorous experiments across three publicly accessible datasets, outperforming other baseline models when aligned with identical backbones.
Additionally, the effectiveness and superiority of our framework are further evidenced through a series of representative visualizations and detailed parameter analysis experiments, solidifying our approach in the field of medical image segmentation.

\section*{Acknowledgments}
This study was partially supported by the National Science Foundation (IIS 2045848), and by the Presidential Research Fellowship (PRF) in the Department of Computer Science at the University of Texas Rio Grande Valley.
Part of the work used Bridges-2 at Pittsburgh Supercomputing Center through bridges2 \cite{brown2021bridges} from the Advanced Cyberinfrastructure Coordination Ecosystem: Services \& Support (ACCESS) program, which is supported by NSF grants \#2138259, \#2138286, \#2138307, \#2137603, and \#2138296.

\bibliographystyle{IEEEtran}
\bibliography{main}

\end{document}